*Accepted to appear in the International Journal of Computer-Supported Collaborative Learning*

# An Artificial Intelligence-driven Learning Analytics Method to Examine the Collaborative Problem-solving Process from a Complex Adaptive Systems Perspective

Fan Ouyang[a], Weiqi Xu[a], Mutlu Cukurova[b]*

[a] College of Education, Zhejiang University, China

[b] UCL Knowledge Lab, University College London, United Kingdom

*Corresponding author e-mail: m.cukurova@ucl.ac.uk

**Abstract:** Collaborative problem solving (CPS) enables student groups to complete learning tasks, construct knowledge, and solve problems. Previous research has argued the importance of examining the complexity of CPS, including its multimodality, dynamics, and synergy from the complex adaptive systems perspective. However, there is limited empirical research examining the adaptive and temporal characteristics of CPS which might lead to an oversimplified representation of the real complexity of the CPS process. To further understand the nature of CPS in online interaction settings, this research collected multimodal process and performance data (i.e., verbal audios, computer screen recordings, concept map data) and proposed a three-layered analytical framework that integrated AI algorithms with learning analytics to analyze the regularity of groups' collaboration patterns. The results detected three types of collaborative patterns in groups, namely the behaviour-oriented collaborative pattern (Type 1) associated with medium-level performance, communication-behaviour-synergistic collaborative pattern (Type 2) associated with high-level performance, and communication-oriented collaborative pattern (Type 3) associated with low-level performance. The research further highlighted the multimodal, dynamic, and synergistic characteristics of groups' collaborative patterns to explain the emergence of an adaptive, self-organizing system during the CPS process. According to the empirical research results, theoretical, pedagogical, and analytical implications were discussed to guide the future research and practice of CPS.

**Keywords:** Collaborative problem solving; Computer-supported collaborative learning; Complex adaptive systems theory; Collaborative pattern; Learning analytics; AI-driven learning analytics

1. Introduction

Grounded upon the social, cultural, and situated perspectives of learning (Vygotsky, 1978), collaborative problem solving (CPS), as one of the computer-supported collaborative learning (CSCL) modes, has been widely used in K-12 and higher education to foster active learning (Hmelo-Silver & DeSimone, 2013; Roschelle & Teasley, 1995; Stahl, 2009). CPS is a *multimodal*, *dynamic*, and *synergistic* phenomenon, where interactive, cognitive, regulative, behavioural, and socio-emotional aspects of collaboration take place synergistically, intertwine inseparably, and influence each other dynamically (Stahl & Hakkarainen, 2021; Vogler et al., 2017). These characteristics of CPS make it compatible with the complex adaptive systems theory, viewing education as a system containing various interdependent elements and dynamic and adaptive interactions between these elements (Byrne & Callaghan, 2014;

Holland, 1996). Drawing upon the complex adaptive systems theory, therefore, traditional educational research methods (e.g., correlational analysis, regression, hierarchical linear modelling) may not be enough to holistically model and monitor the nonlinear and dynamic characteristics of CPS (Amon et al., 2019; Vogler et al., 2017). Analytics of multimodal process data can be more beneficial for understanding the complexity of CPS (Janssen et al., 2013; Medina & Stahl, 2021; Wise et al., 2021). However, there is limited research examining the dynamic and temporal characteristics of CPS, which might lead to an oversimplified representation of the real complexity of the CPS process. To fill this gap, this research proposed a three-layered analytical framework that integrated AI algorithms with learning analytics methods to examine multimodal data collected during groups' CPS processes. This research revealed different types of collaborative patterns of groups, and further examined the associated performance as well as the regularity of each type. Based on the empirical research results, theoretical, pedagogical, and analytical implications were discussed to interpret CPS from a complex adaptive systems theory perspective.

## 2. Literature review

### 2.1 Collaborative problem solving and complex adaptive systems theory

Collaborative problem solving (CPS) requires student groups to solve complex and ill-structured problems without fixed answers to achieve the goal of collective knowledge co-construction (Barron, 2000; Roschelle & Teasley, 1995; Vrzakova et al., 2020). In the past decade, CPS has been widely conducted to promote students' active learning and meaning-making through various learning activities to enable students to connect and construct knowledge in a structured manner and create collective knowledge artefacts (Farrokhnia et al., 2019; Wang et al., 2017). CPS involves multiple levels of dynamic interactions between 1) individual students, 2) individual students and the student group(s), 3) individual students and the learning environment or the knowledge artefact, and 4) the student group(s) and the learning environment or knowledge artefact (Stahl & Hakkarainen, 2021). This multilevel and multilayered nature of CPS indicates a certain level of complexity (Damşa, 2014; Stahl, 2013; Stahl & Hakkarainen, 2021).

Complex adaptive systems theory, generated from the field of biology, emphasizes that a system is a complex entity and its evolution comes from the interactions between microscopic subjects (Byrne & Callaghan, 2014; Holland, 1996). When this concept is applied to the field of education, the teaching and learning processes can be viewed as a complex, self-organizing system, which is the outcome of interdependent interactions between individuals, groups, and teaching and learning components (Amon et al., 2019; Jacobson et al., 2016; Mitchell, 2009) (see Fig. 1). Furthermore, the CPS process is a complex phenomenon where a group of students constantly coordinate interactive, cognitive, regulative, behavioural, and social-emotional aspects through conversations and actions to adapt to the complex and dynamically changing collaborative context, finally to form a self-organizing system. Therefore, the complex system perspective breaks the traditional educational research paradigms such as causal models or linear predictability and highlights the use of organic, nonlinear, and holistic approaches to understanding the nonlinear, dynamic evolution of CSCL (Amon et al., 2019; Holland, 1996).

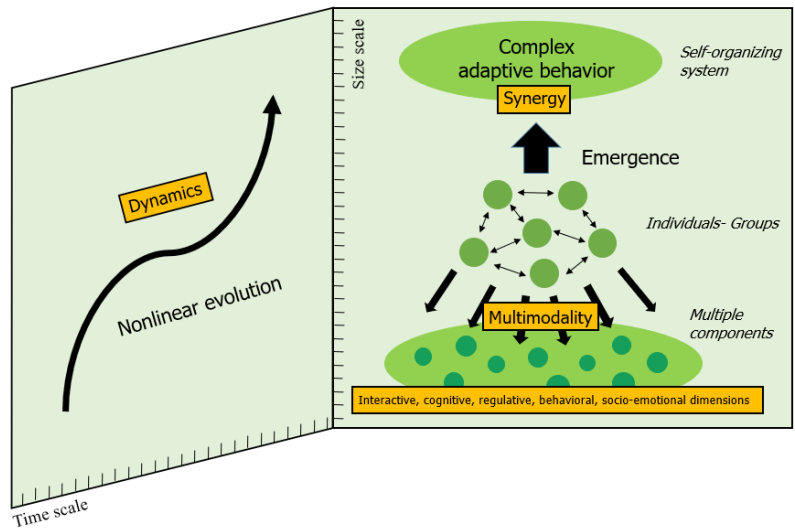

Figure 1. The framework of complex adaptive systems theory (adapted from Holland, 1996)

The conception of complexity is not new to empirical research in the CSCL community. For example, Zuiker et al. (2016) combined complex adaptive systems theory and situated cognition theory to understand and explain collaborative learning as a system-level social activity. Vogler et al. (2017) used complex adaptive systems theory to explore how a message is shared by students in synchronous online discussions and how it triggered students' meaningful engagement in knowledge co-construction. Amon et al. (2019) investigated the system-level team dynamics in collaborative programming from a complex systems theory perspective to describe the dynamic, multimodal, and complex nature of collaboration. From the perspective of complex adaptive systems theory, these studies started to examine and illuminate how, in the dynamic process of CSCL, collaborative meaning-making processes are interwoven with the multimodality and dynamics of individuals, groups, and multiple components during the CSCL process. Moreover, the CSCL community has paid attention to analyzing and understanding CSCL processes from the group or system perspectives, such as collaborative cognitive load, group-level metacognition, or group cognition (Dindar et al., 2020; Kuhn et al., 2020; Zheng et al. 2021). Overall, the complex adaptive systems theory has the potential to provide new perspectives to our understanding of educational phenomena, particularly the collaborative learning process, from a holistic, systematic, and non-linear perspective (Amon et al., 2019; Holland, 1996; Vogler et al., 2017).

**2.2 The characteristics of CPS from the perspective of the complex adaptive systems theory**

Grounded upon the complex adaptive systems theory, group members in the CPS process form an adaptive, self-organizing system that has multi-dimensional characteristics, synergistic relations, and dynamic evolvements (Amon et al. 2019; Barron, 2000; Curşeu et al., 2020; Jacobson et al., 2016; Mitchell, 2009; Stahl & Hakkarainen, 2021). CPS's multimodal characteristics are reflected in the interactive, cognitive, regulative, behavioural, and socio-emotional aspects of the learning process (Fiore et al., 2010; Kwon et al., 2014; Malmberg et al., 2017; Zemel & Koschmann, 2013). On the interactive dimension, the multimodal interaction in CPS is built through students' discourse interaction, text interaction, and behaviour interaction, which is the foundation of collaboration (Ouyang & Xu, 2022; Zemel & Koschmann, 2013). On the cognitive dimension, to solve the ill-structured tasks, students need to share information and propose ideas to co-construct knowledge with group members (Barron, 2000; Ouyang & Chang, 2019; Vrzakova et al., 2020). On the regulative dimension, students need to understand the task, negotiate and plan the goal of the group, and monitor and reflect on the collaborative progress (Malmberg et al., 2017). On the behavioural dimension, students solve problems through online operations and coordinate their behaviours on knowledge artefacts and the behaviours of others (Fiore et al., 2010; Stahl, 2017). On the socio-emotional dimension, students need to build active listening, encourage participation and create cohesive groups to foster active engagement, relaxation of tension, and the emergence of social motivation (Kwon et al., 2014; Rogat & Adams-Wiggins, 2015). More importantly, during the CPS process, various dimensions form complex, interdependent relationships to ultimately contribute to an adaptive, self-organizing system with multilevel, multilayered characteristics (Byrne & Callaghan, 2014; Hilpert & Marchand, 2018).

Furthermore, CPS is a dynamic process, where the relations between elements change over time, and ultimately affect the quality of collaboration (Holland, 1996; Hoppe et al., 2021; Koopmans & Stamovlasis, 2016). During the CPS process, students need to adjust and coordinate interaction, behaviour, cognition, and emotional contributions to the tasks in order to achieve high-quality collaboration reflected in their knowledge constructions and knowledge artefacts (Amon et al., 2019; Barron, 2000; Wiltshire et al., 2019). For example, Amon et al. (2019) found the regularity of coordinative patterns of students' speed of communication, body movement, and screen interaction over time in pair programming can influence their collaborative learning outcomes. Saqr et al. (2021) applied various methods (e.g., process mining, sequence mining map, temporal network) to investigate the relational and temporal dynamics of students' self-regulated learning behaviours during the online collaborative academic writing. Curşeu et al. (2020) proposed a multi-level dynamic model to examine how the variations of core self-evaluations, study engagement, group development, and relationship conflict influenced group identification in collaborative learning. In summary, there is emerging evidence that CPS is a *multimodal*, *dynamic*, and *synergistic* phenomenon, that might be better interpreted through the lens of the complex adaptive systems theory. However, studying CPS through this lens poses significant methodological challenges to traditional statistics and homogenous, unimodal data sources that are frequently used in CSCL research.

## 2.3 The multimodal collaborative learning analytics of CPS

Collaborative learning analytics (CLA), a combination of CSCL and LA, is used to explain, diagnose, and promote collaborative learning processes (Wise et al., 2021). Due to the complexity of CPS, merely focusing on one dimension or perspective of collaboration may cause inconclusive and incomprehensive results. Therefore, CLA advocates the possibility to collect complex and multimodal data (e.g., behavioural, physiological, representational data) to develop comprehensive computational, algorithmic models of collaboration (Blikstein, 2013; Borge & Mercier, 2019; Mu et al., 2020; Ouyang et al., 2022; Sullivan & Keith, 2019). Echoing this trend, recent CSCL research has started utilising multimodal learning analytics (MMLA), integrating multimodal data collection, learning analytics, and AI algorithm-enabled modelling to analyze and mine the complex, dynamic characteristics of CSCL processes (Gorman et al., 2020; Khan, 2017; Olsen et al., 2020; Wiltshire et al., 2019). For example, Khan (2017) proposed a hierarchical computational approach to analyze multimodal data (including audio, video, and activity log files) and modelled the temporal dynamics of student behaviour patterns in collaboration. Wiltshire et al. (2019) used growth curve modelling to examine how students' multiscale movement coordination (e.g., speech, gestures, mouse and keyboard movement) changed across the duration of CPS. Gorman et al. (2020) used computational, quantitative models (including discrete recurrence, non-linear prediction algorithm, and average mutual information) to detect real-time changes in group communication reorganization patterns during collaborative training. The main premise of this emerging research stream is that, compared to traditional learning analytics, such as social network analysis (e.g., Ouyang, 2021) or content analysis (e.g., Jeong, 2013), integration of algorithm-enabled methods in learning analytics and educational data mining can better deal with complex, nonlinear information, extract and represent multi-level and high-dimensional features of CSCL (de Carvalho & Zárate, 2020).

In this research, we present the results of our investigation of groups' collaborative patterns during CPS in an online, synchronous collaboration platform. Taking a complex adaptive systems theory perspective to interpret CPS, we collect and analyze the multimodal process and performance data of student groups, including verbal audios, computer screen recordings, and concept map data. Furthermore, we propose a three-layered analytical framework that integrated learning analytics and AI algorithm-driven methods to examine the collaborative patterns and the regularity of those patterns. We aim to answer two main research questions:

1) *What types of collaborative patterns can be detected from students' multimodal data during their CPS activities in an online synchronous collaboration platform?*
2) *What are the performance differences of groups with different types of collaborative patterns observed during the CPS process in terms of their collaborative products?*

## 3. Methodology

### 3.1 Research context and participants

The research context was two graduate-level seminar courses, titled *Distance and Online Education* and *Online Learning Analytics* offered by the Educational Technology (ET) program at a top research-intensive university in China. The course was designed and facilitated by the same instructor (the second author). She designed a series of CPS activities for small groups (3 or 4 students per group) to work on (see Table 1). Groups were asked to complete one open-ended, ill-structured problem related to the course content in an online, synchronous collaboration platform named *huiyizhuo* (see Fig. 2). The current research dataset consisted of 24 datasets; each dataset included the group's audio recordings of verbal communication data (about 2,160 minutes in total), computer screen recordings of click stream data (about 2,160 minutes in total), text-based chatting, and the final product of concept map data. The instructor occasionally engaged in the CPS activities and her engagement was not the focus of this research; therefore, to maintain data consistency, the instructor data were all removed.

Table 1. Student information

| Group | Participant | Gender | Age | Status | Course |
|---|---|---|---|---|---|
| Group A | A1 | Female | 32 | Graduated student | *Distance and Online Education* |
| | A2 | Male | 41 | Part-time Master student | |
| | A3 | Male | 25 | Potential graduate student | |
| Group B | B1 | Female | 24 | Full-time Master student | *Distance and Online Education* |
| | B2 | Female | 36 | Full-time Ed.D. student | |
| | B3 | Male | 23 | Full-time Master student | |
| Group C | C1 | Female | 24 | Potential graduate student | *Distance and Online Education* |
| | C2 | Male | 31 | Full-time Ph.D. student | |
| | C3 | Female | 27 | Full-time Master student | |
| Group D | D1 | Female | 26 | Full-time Ph.D. student | *Online Learning Analytics* |
| | D2 | Female | 23 | Full-time Master student | |
| | D3 | Female | 23 | Full-time Master student | |
| | D4 | Female | 23 | Full-time Master student | |

The instructional design of CPS activities followed the problem-based learning cycle (Hmelo-silver, 2004): students first analyzed the problem scenario, then identified the knowledge gap needed for solving the problem, and finally generated possible solutions through collaboratively building a concept map. Problems were all open-ended, ill-structured problems that did not have a fixed solution. For example, one CPS activity asked the groups to work as a teaching team to design online learning resources for high school classes. The online collaboration platform *Huiyizhuo* (https://www.huiyizhuo.com/) was used, which provides functions of text chatting, audio and video communication, concept map, note and comment, and resource sharing. In the CPS process, members first communicated through audio and text chatting, then shared resources and kept communications to share knowledge, and finally constructed a concept map to solve problems and propose solutions (see Fig. 2). The concept map was used as the main tool for the collaborative problem-solving process (Novak & Cañas, 2008).

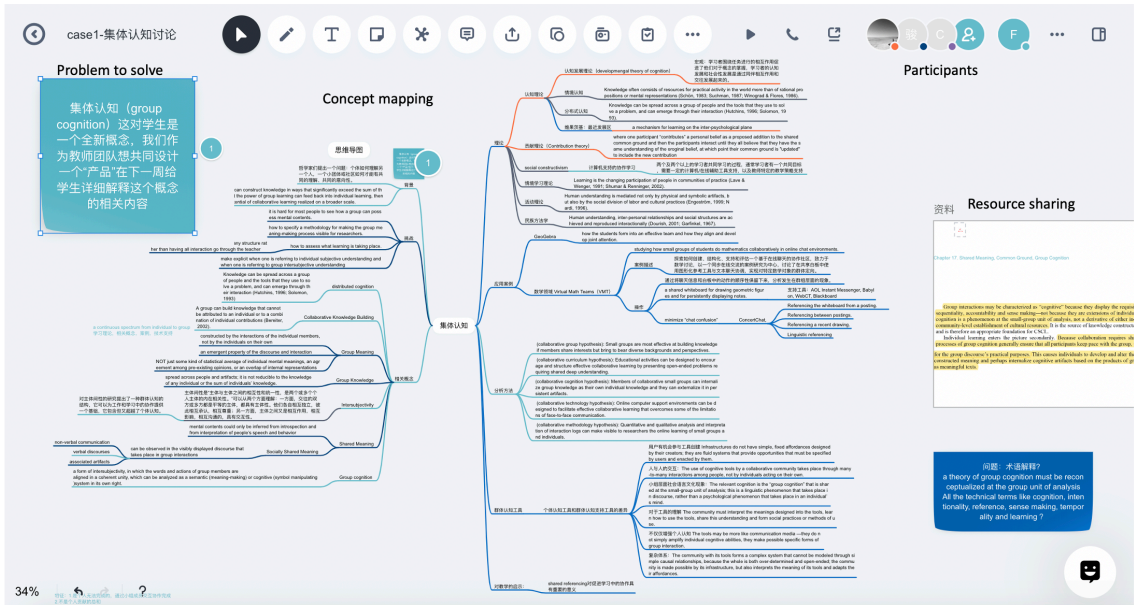

Figure 2. A screenshot of a CPS activity on the *Huiyizhuo* platform

## 3.2 The analytical framework, procedures, and methods

We proposed and used a three-layered analytical framework to examine the characteristics of collaborative patterns during CPS activities (see Fig. 3). AI algorithm approaches were integrated with multiple learning analytics (e.g., quantitative content analysis, clickstream analysis, epistemic network analysis) methods in the analytical framework. In Layer 1 of data pre-processing and analysis, we coded the *interactive, cognitive, behavioural, regulative,* and *socio-emotional* dimensions of CPS (see Table 2). In Layer 2 of multichannel sequence analysis (MCSA), we examined the similarity of groups' CPS sequences to detect distinct clusters of collaboration patterns. In Layer 3, we used a multi-method approach (including statistical analysis, epistemic network analysis, and hidden Markov model) to further examine the regularity among different types of collaboration pattern clusters of student groups.

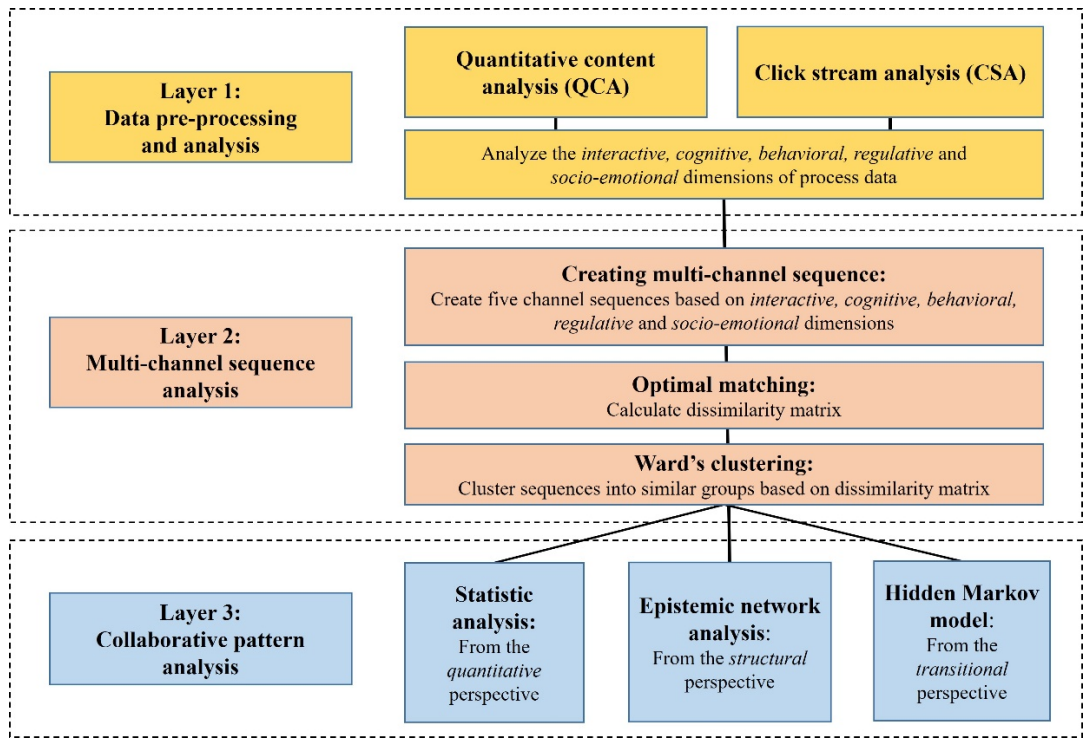

Figure 3. The proposed three-layered analytical framework

**3.2.1 Layer 1: Data pre-processing and analysis**

The computer screen recording data (with audio) was transcribed by two researchers to record participants' verbal communications and online behaviours temporally. During the transcription, the unit was a unit of a sentence (i.e., a full sentence spoken by a student) and a unit of a clickstream behaviour (i.e., a student's mouse clicking or moving operation on the platform). After the transcription, 24 datasets included 11,477 units of verbal and behaviour data in total (Mean = 478.21; SD = 79.18). Drawing from the previous relevant literature, a coding framework was proposed to analyze the group's process data on the *interactive, cognitive, regulative, behavioural,* and *socio-emotional* dimensions (see Table 2). It is worth mentioning that all these five dimensions can be reflected through both verbal communications and online behaviours from the group of students. There are two ways to address this multimodal nature of CPS. One way is assigning students' online behaviours to the cognitive and regulative dimensions (e.g., creating concepts or building on ideas on a concept map can be coded as the cognitive dimension, or moving the mouse to a concept map to observe can be coded as the regulative dimension). In this way, the original behavioural dimension is not recorded (see for instance in Ouyang & Xu, 2022). The second way is to maintain the behaviour as a separate dimension from the cognitive and regulative dimensions. Obviously, there is a tradeoff between these two strategies in terms of dealing with the overlaps of multiple dimensions during the CPS processes. After careful consideration, we decided to adopt the second strategy to retain and demonstrate the analysis results of students' behaviours. This decision is mainly driven by our theoretical positioning of CPS processes as complex and synergistic. The behavioural codes of online behaviours used are not just behaviour-oriented collaboration indicating students work on concept maps, but indirectly involve cognitive and regulative processes (see Table 2). With this strategy, more multimodal and complex characteristics can be captured, compared to the first strategy.

Here we further explain the coding framework that includes the *interactive, cognitive, regulative, behavioural,* and *socio-emotional* dimensions of student groups' CPS processes (see Table 2). The *interactive* dimension recorded students' social interactions through peer communications and online behaviours of the concept map (Ouyang &

Xu, 2022). The *cognitive* dimension analyzed students' knowledge contributions at the superficial, medium, and deep levels (Ouyang & Chang, 2019). The *regulative* dimension represented students' regulation of their collaborative processes, including task understanding, goal setting and planning, and monitoring and reflection (Malmberg et al., 2017). The behavioural dimension analyzed students' online behaviours, including resource management, concept mapping and observation. The *socio-emotional* dimension included active listening and respect, encouraging participation and inclusion, and fostering cohesion during the CPS processes (Rogat & Adams-Wiggins, 2015). It is important to note that although these theory-driven dimensions are indeed useful to label student interactions, they are proxies of tacit learning processes and certain levels of overlap between them are expected. For instance, the behavioural dimension, to a certain extent, might reflect the cognitive (i.e., KS, KM, KD) and regulative (i.e., MR) dimensions (see Table 2). Therefore, the behavioural dimension in this research not merely refers to students' online behaviours, but may also reflect their cognitive and regulative efforts. During the coding process, if one unit included multiple dimensions (e.g., both interactive and socio-emotional dimensions in one unit of verbal communication), we coded with multiple codes in the corresponding unit. When the behavioural unit included the cognitive or regulative dimensions, we not only coded it as the behavioural dimension but also further examined the cognitive and regulative attributes of online behaviours (see 4.1 in the Results section).

Table 2. The coding framework

| Dimension | Code | Descriptions |
| --- | --- | --- |
| Interactive (Ouyang & Xu, 2022) | Peer interaction through communications (Int-C) | A student interacted with peers through verbal communication (i.e., one student responded to others through audio) and texting (i.e., one student replied to others through text) |
| | Peer interaction through behaviours (Int-B) | A student interacted with peers by building on or modifying others' ideas on the concept map |
| Cognitive (Ouyang & Chang, 2019) | Superficial-level knowledge (KS) | A student shared existing information about the topic without further explanations or elaborations |
| | Medium-level knowledge (KM) | A student explained the details of the topic content without further elaborations |
| | Deep-level knowledge (KD) | A student explicitly elaborated the details of the topic content with detailed explanations, support of resources, statistics or personal experiences |
| Behavioural | Resource management (RM) | A student searched or shared resources on the platform or through the Internet |
| | Concept mapping (CM) | A student created, modified, or moved nodes created by himself/herself in the concept map; CM also indirectly reflected students' cognitive processes, including KS (the first-level node created on the concept maps), KM (arguments and explanations added to the second level of the concept map), and KD (examples added to the concept map to further support the arguments as the third level or above) |
| | Observation (OB) | A student moved the mouse over the platform to observe without any operations; OB also indirectly reflected students' regulative dimension (i.e., monitoring and reflection; MR) |
| | Task understanding (TU) | A student read and explained the problems or questions of the tasks |

| Regulative (Malmberg et al., 2017) | Goal setting and planning (GSP) | A student discussed the purpose of the task, divided the task into specific steps, and planned what to do next |
|---|---|---|
| | Monitoring and reflection (MR) | A student monitored the progress of tasks, evaluated the timeline for completing the task, and summarized what had been done and what needed to be done |
| Socio-emotional (Rogat & Adams-Wiggins, 2015) | Active listening and respect (ALR) | A student conveyed attention to other group members by responding to peers after careful listening |
| | Encouraging participation and inclusion (EPI) | A student encouraged the sustained involvement and contributions of group members |
| | Fostering cohesion (FC) | A student conveyed that the group functions as a team (rather than as individuals) by working together, referring to the group as "we" |

*Note.* The coding framework clearly describes the codes we used as well as cites the previous literature from which the codes were derived.

Three raters completed the coding procedure. Rater 1 first coded 30% of the dataset based on the proposed coding scheme. Next, rater 2 coded the data again and had multiple meetings with rater 1 to solve discrepancies. Krippendorff's (2004) alpha reliability was 0.802 among two raters at this phase. Finally, rater 1 completed the coding of the rest of the dataset, then rater 3 double-checked the coding results, and consulted with rater 1 to decide the final codes if there were any conflicts.

### 3.2.2 Layer 2: Multichannel sequence analysis

After data pre-processing and analysis, we applied multichannel sequence analysis (MCSA) to examine the similarity of groups' CPS activities to detect types of collaborative patterns. MCSA, as a sequence analysis method generated from the field of bioinformatics, is used to analyze multiple parallel trajectories (e.g., dimension, status) of time sequences simultaneously (Gauthier et al., 2010).

There were three steps for MCSA. In Step 1, the five-dimensional codes were transformed into five-channel sequences, and 24 five-channel sequences were created in total. Because the CPS activities occurred in an authentic, natural collaborative context, the lengths of the 24 five-channel sequences were different. We decided not to standardize the length of sequences in order to keep original information.

In Step 2, the optimal matching (OM) algorithm was used to calculate and align 24 five-channel CPS sequences to detect the similarity of groups' collaboration patterns. Based on the Levenshtein distance calculation, the OM algorithm computed the similarity between the corresponding channels of each pair of sequences in terms of edit distances (Abbott & Tsay, 2000). Edit distances were the transformation steps (insertion, deletion, and substitution), that convert from one sequence to another. The similarities between sequences were determined according to the generalized Hamming distance with user-defined transformation costs. The costs for insertion, deletion, and substitution were set to be the same (i.e., 1) in all dimensions to give them equal weight. No cost was set up for substituting missing states since we only detected similarities based on the observed trajectories. OM was run through R packages *TraMineR* (Gabadinho et al., 2011) and *seqHMM* (Helske & Helske, 2019).

In Step 3, Ward's clustering (WC) algorithm was used to cluster CPS activities into types with similar collaborative patterns. WC is a hierarchical bottom-up algorithm that computes similarities between sequences and clusters them into similarity-based groups (Murtagh & Legendre, 2014). WC was chosen because comparing to other clustering methods, it produces usable and relatively even-sized clusters. The choice of clusters was based on goodness-of-fit statistics, the dendrogram, and the interpretability of the clusters. WC was carried out through the R package *cluster* (Maechler et al., 2015).

### 3.2.3 Layer 3: Collaborative pattern analysis

In Layer 3, three analytics methods were used to reveal the quantitative, structural, and transitional characteristics of the collaborative patterns identified in step 2. First, from a quantitative perspective, statistical analysis (SA) was used to analyze the frequency of interactive, cognitive, regulative, behavioural and socio-emotional dimensions and then a one-way analysis of variance (ANOVA) was conducted to test the significance of differences among clusters.

Second, from a structural perspective, epistemic network analysis (ENA) was performed to demonstrate the accumulative connections of the interactive, cognitive, regulative, behavioural, and socio-emotional dimensions among different collaboration pattern types. ENA can identify and quantify connections among elements in coded data and represent them in dynamic network models (Shaffer et al., 2016). An ENA Webkit (epistemicnetwork.org) was used to perform ENA analysis and its visualization (Marquart et al., 2018). However, the structure did not merely focus on the cognitive dimension but included all five dimensions.

Third, from a transitional perspective, an algorithm model named hidden Markov model (HMM) was used to describe a Markov Chain with implicit unknown parameters, detect a latent process with several hidden states, and find transitional modes between hidden states (Eddy, 1996). Compared to other algorithms for dynamic analysis (e.g., dynamic Bayesian network), the strength of HMM is compressing the complex sequence data into hidden states (i.e., in our case the collaboration stages) as well as capturing the dynamic changes between states, which cannot be represented by the observed sequences (Eddy, 1996; Felsenstein & Churchill, 1996). R package *seqHMM* (Helske & Helske, 2019) was used to analyze HMM based on clusters of sequences. A separate HMM was created and fitted for each type of collaboration pattern detected. The dataset included 15,750 codes for Type 1, 32,130 codes for Type 2, and 13,000 codes for Type 3. For each type, we pre-specified the number of states in HMM in a range from 2 to 9 to find the best model. The expectation-maximization (EM) algorithm was used to estimate parameters and fit HMMs models. In order to reduce the risk of being trapped in a poor local optimum, the HMMs were run for 100 iterations with random starting values set by R package *seqHMM*. Bayesian information criterion (BIC) was used to choose the optimal number of hidden states in each HMM. Specifically, the lower value of BIC is chosen to assess the comparative model fit.

Finally, a previously-validated assessment framework (Novak & Cañas, 2008) was utilized to evaluate each group's concept map data as the final performance (see Table 3). The assessment included three dimensions, i.e., proposition, hierarchy, and example. Proposition refers to the basic themes and concepts about the topic (a meaningful proposition was scored 1 point); hierarchy represents the certain hierarchical structure of a concept map (an effective hierarchy was scored 5 points); and the example is used to reflect the corresponding themes and concepts (a valid example was scored 1 point). The final score of a concept map was the sum of scores of three dimensions. Two raters evaluated each concept map individually and reached an interrater reliability of Krippendorff's (2004) alpha value of 0.912. After consulting with the first author, the final scores were confirmed.

Table 3. Assessment of concept map (adapted from Novak & Canas, 2008)

| Dimension | Description | Scoring rules |
| --- | --- | --- |
| Proposition | Did the concept map reflect basic themes and concepts? Was the relationship between topics correct and appropriate? | Each meaningful idea, concept, or argumentation was scored 1 point |
| Hierarchy | Did the concept map show a certain hierarchy? Was each subordinate concept more specific than the previous one? | Each effective hierarchical structure was scored 5 points |
| Example | Did examples in concept maps reflect their corresponding themes, concepts, or labels? Were examples used effectively and properly? | Each appropriate example or evidence was scored 1 point |

## 4. Results

After Layer 1 of pre-processing and analysis, five dimensions (i.e., interactive, cognitive, regulative, behavioural, and socio-emotional) were coded in each CPS activity and transformed into 24 five-channel sequences. The optimal clustering results generated from Layer 2 revealed *three* types of collaborative patterns, consisting of 5, 14, and 5 CPS activities for Type 1, Type 2, and Type 3, respectively (see Fig. 4 and Fig. 5). As we can see from Figure 4, 24 datasets were generated from four groups (Groups A, B, C, and D), the distribution of four groups in the three types was relatively equal. Regarding the final performances of collaborative concept maps, although there was no statistical difference identified with the ANOVA tests, Type 2 had the highest score of concept maps (Mean = 105.29, SD = 33.64), followed by Type 1 (Mean = 92.60, SD = 23.67), and Type 3 (Mean = 77.60, SD = 28.18). In summary, Type 1 was associated with medium-level performance, Type 2 was associated with high-level performance, and Type 3 was associated with low-level performance.

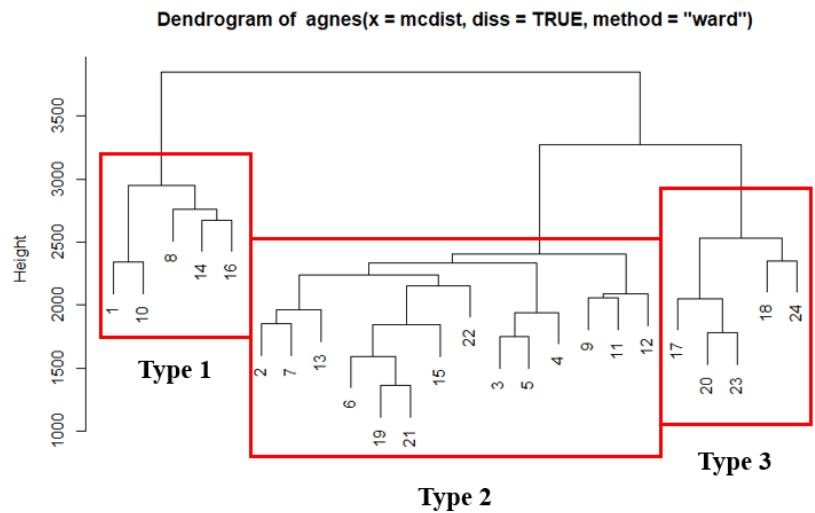

Figure 4. Ward's clustering results for 24 CPS sequences

*Note.* In Type 1, no. 1 was Group D, no. 8 and 14 was Group B, no. 10 and 16 was Group A; In Type 2, no. 2, 3, 4, 5 were Group D, no. 6, 11 were Group B, no. 7, 9, 12, 15, 21 was Group C; no.13, 19, 22 was Group A; In Type 3, no. 17, 20, 23 was Group B, no. 18, 24 was Group C.

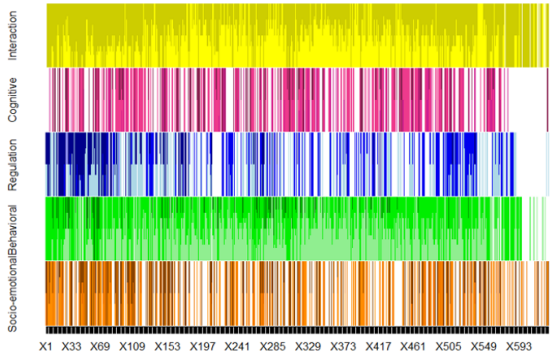
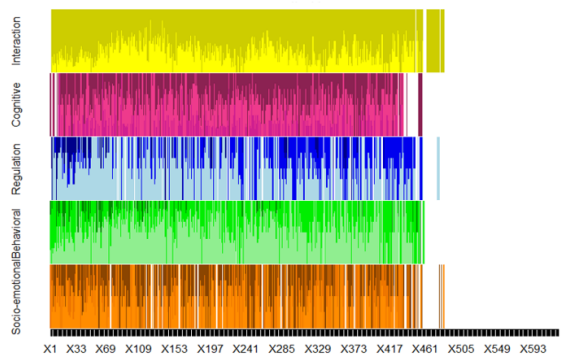
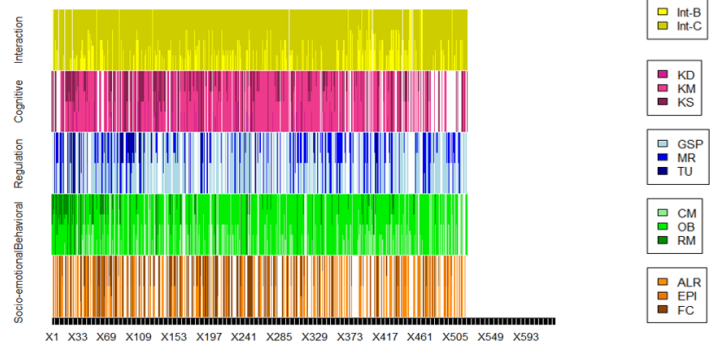
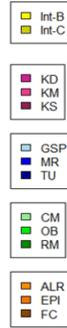

Figure 5. Three types of collaborative patterns from optimal clustering results

*Note.* Each channel represents how the codes of a dimension change in time series. Specifically, the first channel (yellow) represents the *interactive* dimension (Int-C and Int-B); the second channel (pink) represents the *cognitive* dimension (KS, KM, and KD); the third channel (blue) represents the *regulative* dimension (GSP, MR and TU); the fourth channel (green) represents the *behavioural* dimension (CM, OB, and RM); the fifth channel (orange) represents the *socio-emotional* dimension (ALR, EPI, and FC). The blank part indicates that no code was generated.

### 4.1 From a quantitative perspective

ANOVA with the Bonferroni correction was conducted to test the significant differences between the three types on the five dimensions. Before ANOVAs, Levene tests were conducted; and the results showed the homogeneity of variance. Moreover, a series of post-hoc pairwise comparisons were conducted to further reveal significant differences between types (see Table 4). Considering that some data (e.g., TU, RM, OB, and EPI) were not normally distributed, a non-parametric test was conducted to cross-check the ANOVA results. The results showed that there were significant differences in the frequency of Int-B, KS, KM, TU, RM, CM, and OB ($p < 0.05$) with the Bonferroni correction under the Kruskal-Wallis test. Specifically, there were statistically significant differences between the three types on the *behavioural* dimension (RM, CM, OB). Regarding the main behaviours of concept mapping (i.e., CM), Type 1 ranked first among the three types, followed by Type 2 and Type 3. CM indirectly reflected students' *cognitive* engagement (i.e., KS, KM, and KD) when creating and editing nodes in the concept maps (Type 1: KS: Mean = 9.40, KM: Mean = 18.80, KD: Mean = 61.40; Type 2: KS: Mean = 7.64, KM: Mean = 16.07, KD: Mean = 51.14; Type 3: KS: Mean = 4.00, KM: Mean = 9.60, KD: Mean = 17.20). In Type 3, more OB behaviors (Mean = 146.8, SD = 53.38) were detected than interactive behaviors (RM: Mean = 20.6, SD = 13.70;

CM: Mean = 42.2, SD = 7.19). Moreover, OB, to some extent, also reflected students' monitoring of others' operations in the *regulative* dimension. Moreover, on the *interactive* dimension, statistical significance was found on Int-B, where Type 1 had the highest frequency, followed by Type 2 and Type 3; but no statistical significance was found on Int-C, where all three types had a high level of Int-C. On the *cognitive* dimension, statistical significances were found on KS (Type 2 > Type 1) and KM (Type 3 > Type 1, Type 3 > Type 2), except KD (all three types had a low level of KD). On the *regulative* dimension, statistical significance was found on TU (Type 1 > Type 2, Type 1 > Type 3), while no statistical significance was found on GSP and MR. In addition, no statistical significance was found in the *socio-emotional* dimension; there was a low level of frequency in the three codes under the socio-emotional dimension. We concluded that there were different collaborative patterns among the three types.

Table 4. Results of frequencies and one-way ANOVAs of the three types

| Code | Type 1 (n=5) Mean (SD) | Type 2 (n=14) Mean (SD) | Type 3 (n=5) Mean (SD) | ANOVA F | p | Post-hoc pairwise comparison |
|---|---|---|---|---|---|---|
| Interactive | | | | | | |
| Int-C | 215.20 (103.38) | 213.93 (67.91) | 239.60 (60.72) | 0.23 | > .10 | |
| Int-B | 176.20 ( 19.83) | 105.43 (42.95) | 52.40 (20.98) | 14.91 | < .001*** | Type 1 > Type 2 > Type 3 |
| Cognitive | | | | | | |
| KS | 23.60 (17.95) | 56.36 (21.68) | 39.60 (13.72) | 5.43 | < .05 ** | Type 2 > Type 1 |
| KM | 54.20 (32.51) | 59.14 (27.25) | 105.00 (29.73) | 5.36 | < .05 ** | Type 3 > Type 1, Type 3 > Type 2 |
| KD | 17.40 (12.74) | 20.64 (16.09) | 11.60 ( 5.41) | 0.77 | > .10 | |
| Regulative | | | | | | |
| TU | 28.80 (12.76) | 7.36 ( 6.34) | 11.00 ( 6.36) | 13.44 | < .001*** | Type 1 > Type 2, Type 1 > Type 3 |
| GSP | 49.60 (21.95) | 46.71 (23.88) | 70.40 (27.32) | 1.80 | > .10 | |
| MR | 39.80 (18.90) | 24.50 (17.30) | 31.40 (10.43) | 1.63 | > .10 | |
| Behavioural | | | | | | |
| RM | 20.20 (17.66) | 6.86 ( 7.77) | 20.60 (13.70) | 4.04 | < .05 ** | Type 1 > Type 2, Type 3 > Type 2 |
| CM | 160.80 (37.31) | 97.36 (34.18) | 42.20 ( 7.19) | 17.67 | < .001*** | Type 1 > Type 2 > Type 3 |
| OB | 154.20 (75.62) | 54.29 (26.77) | 146.80 (58.38) | 12.41 | < .001*** | Type 1 > Type 2, Type 3 > Type 2 |
| Socio-emotional | | | | | | |

| | | | | | |
|---|---|---|---|---|---|
| ALR | 35.00 | 36.43 | 35.00 | 0.04 | > .10 |
| | (11.64) | (13.30) | (13.51) | | |
| EPI | 24.80 | 19.64 | 10.80 | 0.94 | > .10 |
| | (24.66) | (15.55) | ( 5.45) | | |
| FC | 36.40 | 28.29 | 32.40 | 0.30 | > .10 |
| | (23.42) | (20.41) | (18.80) | | |

*Note.* \* $p < .10$, \*\* $p < .05$, \*\*\* $p < .001$.

For simplicity, although the cognitive and regulative dimensions were reflected by online behaviours, the details of the results related to cognitive (KS, KM, and KD) as well as regulative (MR) dimensions were only described in the text and not shown in Table 4.

**4.2 From a structural perspective**

From a structural perspective, the regular characteristics among the three types were reflected by the connection values and the locations of the centroid of the ENA plots (see Fig. 6). First, regular characteristics were reflected by three pairs of connected codes (connection values > 0.40) among the three types. They were Int-C – KM, Int-C – OB, and Int-C – Int-B. The *socio-emotional* dimension was weakly associated with other codes (connection values < 0.40) in the three types.

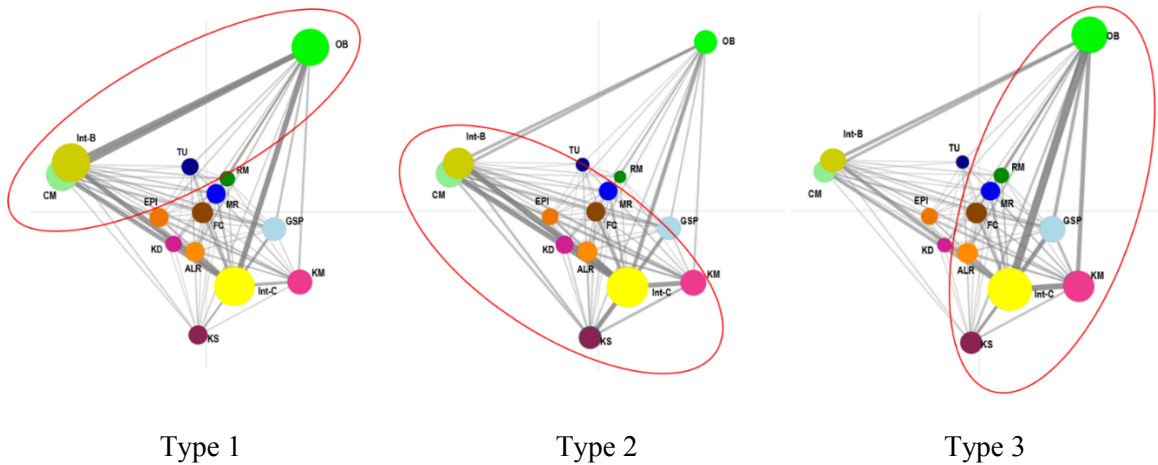

Type 1            Type 2            Type 3

Figure 6. The epistemic network analysis of the three types of collaboration patterns

*Note.* The colours of the codes are not automatically generated through ENA Webkit, but set manually by researchers according to the code colour.

Different characteristics were identified among the three types of collaboration pattern clusters, reflected by the locations of the centroid in epistemic networks (shown as red circles in Fig. 6). In Type 1, the centroid of the epistemic network was located at the upper left corner, focusing on the behaviour-related codes (including OB, CM, and Int-B). The connection between Int-B and OB was 0.74, the connection between Int-B and CM was 0.72, and the connection between OB and CM was 0.69. In Type 2, the centroid was located at the bottom of the epistemic network, focusing on the behaviour and communication-related codes (including CM, KM, KS, Int-B, and Int-C). The connection between Int-B and Int-C was 0.71, the connection between Int-C and CM was 0.67, the connection between Int-C and KM was 0.62, the connection between Int-C and KS was 0.61, and the connection between Int-B and CM was 0.60. In Type 3, the centroid of the epistemic network was located on the right side, mainly focusing on communication-related codes (including KM, GSP, and Int-C). The connection between Int-C and OB was 0.90, the connection between Int-C and KM was 0.73, and the connection between Int-C and GSP was 0.59. In summary,

Type 1 concentrated on the *behaviour-related* codes; Type 2 focused on both *communication-related* and *behaviour-related* codes, and Type 3 concentrated on the *communication-related* codes.

**4.3 From a transitional perspective**

From a transitional perspective, similar characteristics among the three types of collaboration pattern clusters were observed in HMM results. Specifically, a 5-state HMM of Type 1, a 7-state HMM of Type 2, and a 5-state HMM of Type 3 were selected for the best model fits with the lowest values in BIC (see Table 5, Fig. 7, and Fig. 8). Among the HMM results, the regular characteristics of transitional mode began by *communication* (State 1 in Type 1, 2, 3), then moved to *behaviour* (State 2 in Type 1, 3 and State 2, 4 in Type 2), and finally returned to *communication* (State 3, 4 in Type 1, 3 and State 5, 6 in Type 2) (see Table 5 and Fig. 7).

Table 5. Descriptions of hidden states in HMMs of the three types

| Hidden state | Type 1 | Type 2 | Type 3 |
| --- | --- | --- | --- |
| 1 | Students communicated with each other | Students interacted with each other through communication and behaviours | Students communicated with each other and sometimes observed |
| 2 | Students operated the concept map together and sometimes observed | Students operated the concept map together while sometimes observing others or communicating with each other | Students observed others, sometimes communicated with each other, operated the concept map, or managed resources |
| 3 | Students constructed knowledge through peer communications | Students communicated with each other to set goals and plans | Students constructed knowledge through peer communications |
| 4 | Students mainly operated the concept map and sometimes communicated with each other | Students operated and modified the concept map and sometimes observed others | Students communicated with each other and sometimes observed |
| 5 | Collaboration ended | Students communicated with each other | Collaboration ended |
| 6 | | Students interacted with each other through communication and behaviours | |
| 7 | | Collaboration ended | |

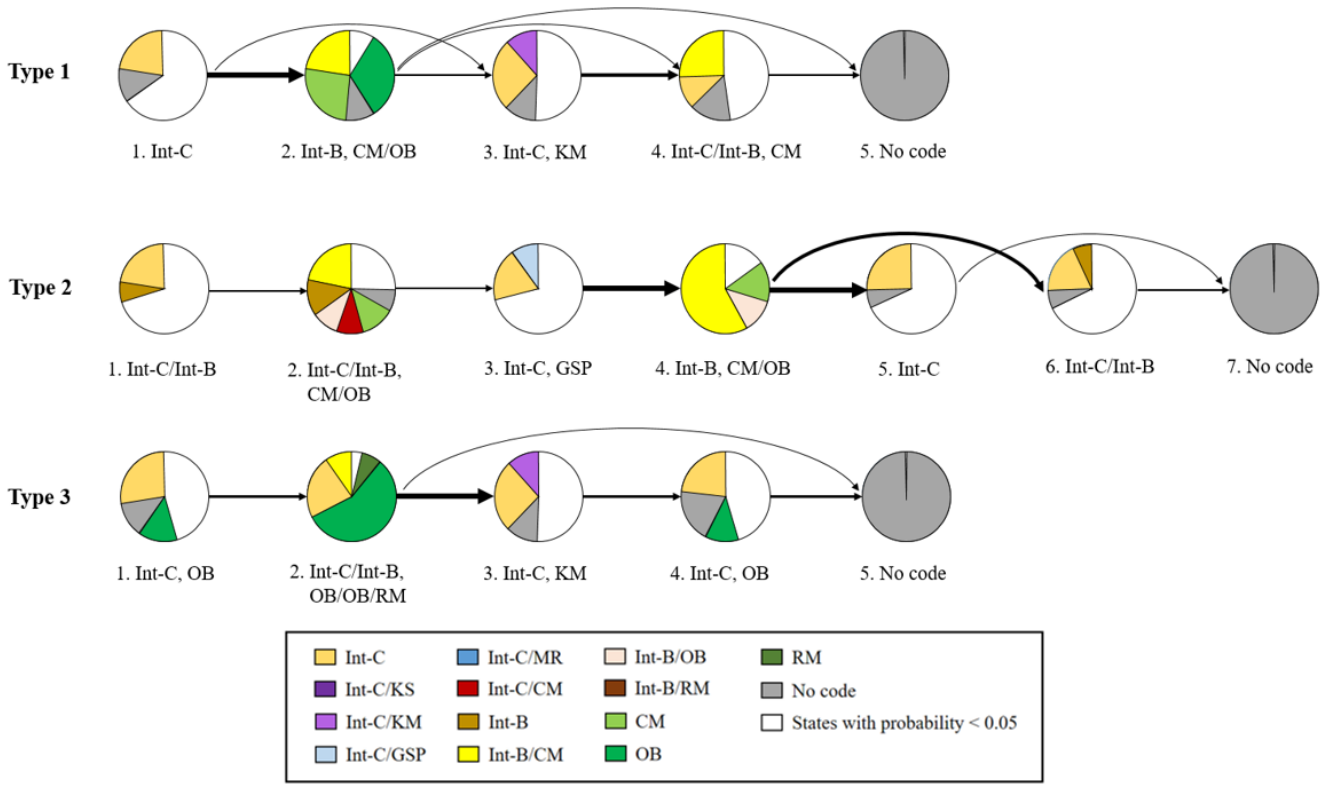

Figure 7. HMM graphs of the transitional structures among the three types

*Note*. HMMs are shown as directed graphs, where the pies represent hidden states and the slices show the probabilities of observed states within each hidden state. Observed states are shown as labels, which represent the codes observed at the same time from five dimensions (probabilities > 0.05 is shown). The arrows indicate transition probabilities between the hidden states; the thicker the arrow, the higher the probability. No code is detected when no code is generated in all five dimensions.

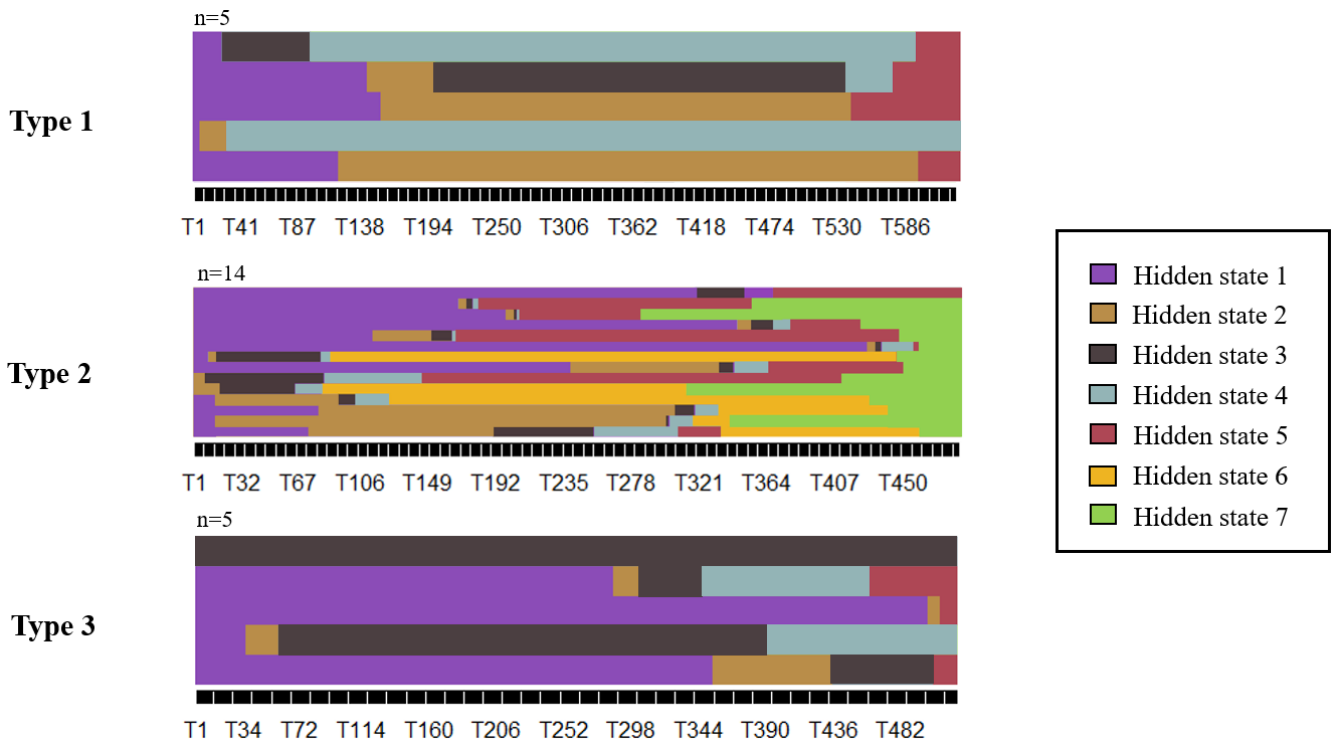

Figure 8. Most probable HMM paths among the three types

Regularity of the transitional characteristics was detected, where Type 1 was characterized as a *behaviour-oriented transition*, Type 2 was characterized as a *communication-behaviour-synergistic transition*, and Type 3 was characterized as *a communication-oriented transition*. In Type 1, groups were more likely to transition from other states to the behaviour-related states (State 2, 4) (see Fig. 7). Additionally, states related to peer communications (State 1, 3) were short and infrequent, while states of peer behaviours (State 2, 4) were long and frequent (see Fig. 8). In Type 2, students had a high probability of transitioning to State 4 and State 6, which included both communication-related and behaviour-related codes. Except for State 3 and State 4, each state in Type 2 shared a balance of time and frequency across the CPS processes. In Type 3, groups had the highest probability of transitioning to a communication-related state (State 3). States dominated by peer communications (State 1, 3) were long and frequent, while states dominated by peer behaviours (State 2, 4) were short and infrequent.

## 5. Discussion

### 5.1 Addressing research questions

Due to the complexity of CPS (Amon et al. 2019; Jacobson et al., 2016; Stahl & Hakkarainen, 2021), unimodal data and traditional statistics may be limited for holistically modelling it (Ouyang et al., 2022). This research collects multimodal data, including verbal audios, computer screen recordings, and concept map data, and proposes a three-layered analytical framework integrating learning analytics and AI algorithm-driven methods to investigate the collaboration patterns of groups working through an online collaboration platform. Compared to previous studies, this approach proposes a multilayered approach to better understand the complex nature of CPS from an organic, nonlinear, and holistic perspective that is aligned with the complex adaptive systems theory (Vogler et al., 2017; Zuiker et al., 2016). To answer the research questions, three types of collaborative patterns were detected with different levels of the final concept map performance: Type 1 was characterized as a behaviour-oriented collaborative pattern and associated with medium-level performance, Type 2 was characterized as a communication-behaviour-synergistic collaborative pattern and associated with high-level performance, and Type 3 was characterized as a communication-oriented collaborative pattern and associated with a low-level performance. More importantly, the attributes of the three types were together demonstrated through the quantitative frequency, structural, and transitional aspects as the results showed. Regarding the behaviour of concept mapping (i.e., CM), Type 1 ranked first among the three types, followed by Type 2, and Type 3. Since the complexity of CPS, the cognitive dimension was also reflected through students' online behaviours as knowledge sharing and construction during the concept mapping process, which partially explained the relations between the groups' behaviours and final performances. For example, Type 3 had the lowest level of cognitive engagement (KS, KM, and KD) through online behaviours, which also resulted in the lowest score of the final performance among the three types. The ENA and HMM results also indicated different characteristics of the three types. For example, Type 2 shared strong connections between communication-related and behaviour-related codes with a communication-behaviour-synergistic transition; Type 1 generated strong connections only within the behaviour-related codes with a behaviour-oriented transition; Type 3 generated strong connections only within the communication-related codes with a communication-oriented transition. Overall, this research revealed three types of collaborative patterns in CPS as well as their characteristics to further verify the multimodality, dynamics, and synergy of the CPS process.

## 5.2 Theoretical implications: The complexity of collaboration

The complexity theory, as an emerging paradigm in educational research (Morrison, 2002), looks at and examines the educational system in ways which break with the simple cause-and-effect models, linear modelling, and regression and replaces them with organic, non-linear, and holistic approaches (Cohen et al., 2013). The recently published international handbook of computer-supported collaborative learning calls for the application of AI algorithm-based models to model complex systems and learner self-organization for collaboration at various scales (Cress et al., 2021). From the perspective of complex adaptive systems theory, CPS is a complex phenomenon, in which group members interact with each other, the learning environment, or the knowledge artefacts adaptively to form a group-level, collaborative pattern and collective intelligence (Jacobson et al., 2016). This research provides a new method to explain the emergence of a self-organizing system during the CPS process by integrating AI algorithms with learning analytics and various modalities of data to analyze *multimodal*, *dynamic*, and *synergistic* characteristics of collaboration.

First, the results demonstrated the *multimodal* characteristics of collaborative learning, reflected by the interactive, cognitive, regulative, behavioural, and socio-emotional dimensions, that formed the foundation for the emergence of an adaptive, self-organizing system as observed in Markov models' distinct states of different collaborative patterns. Consistent with previous studies (e.g., Ouyang & Chang, 2019; Park et al., 2015), we found that groups' peer interactions were primarily reflected in students' verbal communications, and the cognitive and regulative dimensions were closely related to peer interactions via verbal discourses. In other words, active peer communications formed a foundation for deep-level knowledge construction and group regulation (Zemel & Koschmann, 2013). On the behavioural dimension, collaborative patterns with high and medium performance had more active behaviours (i.e., concept mapping, and resource management), while low performing patterns had more passive behaviours (i.e., observation). In addition, the cognitive dimension was also reflected in students' online behaviours as knowledge construction in concept mapping, which might partially explain the relations between groups' behaviours and final performance. For example, our results showed that Type 3 had the lowest level of knowledge construction through students' online behaviours, which also resulted in the lowest collaborative performance among the three types. Moreover, more deep-level knowledge was generated through concept mapping behaviours than verbal communications. Therefore, in the multimodal CPS process, we could infer that online behaviour might play a critical role in transforming students' superficial-level and medium-level knowledge in verbal discourse into a deep-level knowledge co-construction (Ouyang & Xu, 2022). However, inconsistent with previous studies that highlighted the role of social emotions (e.g., Kwon et al., 2014; Rogat & Adams-Wiggins, 2015), no significant difference was found between collaborative patterns on the social-emotional dimension. Besides, the socio-emotional dimension was weakly-connected with other dimensions. One of the potential reasons is that the traditional cultural settings of the East make students less likely to express their social emotions during collaboration (Zhang, 2007; Zhang, 2013). These results indicate the complex connections between multiple dimensions emerged into the multimodality of CPS which in turn influences the collaborative outcomes of group interactions.

Second, the *dynamic* characteristics uncovered how an adaptive, self-organizing systems emerged within groups during the CPS processes as observed in Markov models' distinct states for each type of patterns. When confronted with a new problem, groups tended to discuss it together to understand the task and set a specific goal (i.e., communication state). Initial verbal communications in this stage helped group students adapt to the social, collaborative environment and become familiar with the group members, which is beneficial to transferring students to active collaborators (Kwon et al., 2014). However, less cognitive engagement was found in the initial communication stage. Since meaningful knowledge construction belongs to a relatively higher level in CPS, it might require more complex factors (Roschelle & Teasley, 1995), it might also appear based on enough peer interaction and the feeling of safety and collaboration (Schindler & Bakker, 2020; Zemel & Koschmann, 2013). After that, some students started operating concept maps or managing resources, and others kept observing (i.e., behaviour state). In this stage, students' cognitive engagement and metacognitive regulation are mainly reflected through their online behaviours and sometimes presented by means of verbal communication. When new difficulties or bottlenecks appeared, students would suspend their work to discuss it again and reflect on how to solve them (i.e., communication state). This research demonstrated how groups evolved through the developmental stages (i.e., from communication to behaviour and back to communication) to adapt to the ongoing changes in collaboration demands and finally form a self-organizing system with relatively stable states and their associated transition probabilities.

Third, as one of the keys to succeed in forming an adaptive, self-organizing system during collaboration, the *synergistic* characteristics were embodied in groups' coordination between communications and behaviours. Similar to previous studies (e.g., Amon et al., 2019; Ramenzoni et al., 2012; Wiltshire et al., 2019), these findings indicated that high-performance groups (i.e., Type 2) gave rise to synergistic collaboration across verbal discourses and online behaviours in order to adapt to the shifting task demands. However, considering the external factors (e.g., task difficulty, group configuration, social learning context), it is challenging for all groups to collaborate synergistically and to achieve a high quality of collaboration. For example, the medium-performing groups (i.e., Type 1) were more likely to involve online behaviours (e.g., concept mapping) to drive CPS activities, while the low-performing groups (i.e., Type 3) were more likely to conduct CPS activities through verbal discourses. Furthermore, the ENA connection values were relatively stronger in Type 1 and Type 3 than in Type 2, which to some extent verified the difficulty of achieving a high-quality synergy between communication and behaviour in the CPS processes. Overall, this research highlights the complex characteristics of collaborative learning, which further argues for multi-dimensional, multi-level, and multi-perspectival analytical approaches to collaborative learning research (Cohen et al., 2013; Jacobson et al., 2016; Morrison, 2002). This research builds a bridge between the complex adaptive systems theory and analytics of collaborative learning to guide future empirical studies in the CSCL field that would take similar approaches to the investigation of students' group interactions.

### 5.3 Analytical implications

Since collaborative learning is a complex, adaptive process, this research extends investigations on using AI-driven learning analytics to understand *multimodal*, *dynamic*, and *synergistic* characteristics of groups' collaborative patterns during a complex collaborative task. On the one hand, multimodal data collection and analysis are suggested for the investigation of complex educational phenomena and problems (Cukurova et al., 2020; Michail Giannakos, Roberto Martinez-Maldonado; Stahl & Hakkarainen, 2021; Vogler et al., 2017). Recent research has used multimodal data (e.g., speed rate, gesture, body movement, eye movement) to examine the collaboration patterns and characteristics in CSCL, which can complement analysis results from traditional discourse, online content, and product data (Amon et al., 2019; Blikstein, 2013; Mu et al., 2020). More importantly, due to the complexity of CPS, the multiple dimensions (e.g., interactive, cognitive, regulative) might be reflected through communication and the behaviour of collaboration. For example, the behaviours in this research also indirectly reflected the cognitive engagement and metacognitive regulation of the groups of students during the CPS processes. On the other hand, this research verified that the AI algorithm-enabled learning analytics and data mining approach can help to reveal complex characteristics of CPS from quantitative, structural, and transitional perspectives. Compared to traditional learning analytics methods, the integration of learning analytics approaches and algorithm-enabled methods can better process multimodal and nonlinear data in order to extract and represent the complex and dynamic structure of CSCL (de Carvalho & Zárate, 2020). For example, HMM has the potential to capture the stable, distinct states of CPS as well as dynamic movements between and within them, which can particularly reveal the adaptive self-organizing system attribute of collaborative learning. Furthermore, advanced AI algorithms (e.g., multidimensional recurrence quantification analysis, natural language processing, genetic programming) have been applied in CSCL to analyze and reveal the complexity and dynamics of collaboration (Amon et al., 2019; de Carvalho & Zárate, 2020; Hoppe et al., 2021; Sullivan & Keith, 2019). Future work can further integrate these advanced AI algorithms with learning analytics and data mining to reveal the multilevel, multidimensional characteristics of CSCL. There is a call for the application of technological advances (such as machine learning) to foster real-time feedback and assessment of group work, identify collaboration patterns to effectively guide productive collaboration, and accurately predict collaborative outcomes (Amon et al., 2019; Eloy et al., 2019).

## 5.4 Pedagogical implications

CPS is a complex process that is unlikely to follow a particular model of linear sequences of actions. Therefore, during the CPS process, instructors should provide varied scaffoldings as well as flexible support to promote students' collaboration. First, instructors can provide different scaffoldings to enhance groups' collaboration quality based on the multimodal characteristics of CPS. For example, the foundational role of active communications was reinforced by the research results, therefore, instructors can regulate groups' communications through metacognitive scaffoldings (e.g., monitor and stimulate student tasks) to foster students' cognitive and regulative engagement (Ouyang et al., 2021; Van Leeuwen et al., 2015). In addition, due to the weak connection with the socio-emotional dimension as indicated above, the instructor's social support, such as encouragement and affirmation, might ease the atmosphere and mobilize students' social emotions (Ouyang & Scharber, 2017; Ouyang et al., 2020). Second, instructors are supposed to provide flexible and changeable support (Park et al., 2015; Van de Pol et al., 2010) in terms of student groups' dynamic evolvement in CPS (i.e., from communication to behaviour and back to communication) we detected in the current study. To be specific, in the initial stage, instructors can regulate students to organize discussion about the topic; in the middle stage of online operation, instructors can reduce intervention, observe students and provide certain support when the group needs; in the latter stage, instructors can guide students to solve problems through information or content on the topic (Kaendler et al., 2015; Ouyang & Xu, 2022; Van Leeuwen et al., 2015). Third, it is challenging to achieve the high-quality synergy between communication and behaviours of student groups, instructors should pay attention to groups' inactive moments in verbal communication or online behaviour to foster synergistic co-construction of meanings (Barron, 2000, Brown et al., 1989; O'Donnell & Hmelo-Silver, 2013). When students only focus on the discussion and neglect operating the task, the instructor can remind them to return to online operations to drive the task; when students lack enough communication, instructors can guide them to discuss the topics together. Taken together, from a pedagogical perspective, instructors should be aware of groups' dynamic and synergistic status, and support their work appropriately with varied scaffoldings to promote students' CPS.

**5.5 Limitations and future directions**

There were four major limitations in the current research. The first limitation of this research is the sample size of student groups (4 groups generated 24 collaborative activities), with a limited range of demographic backgrounds. The small sample size might be a reason for causing the insignificant differences in the final concept map products. Therefore, future empirical research needs to expand the sample size and experiment with different courses to test, validate, or modify the implications. The second limitation is that we do not control the instructor's participation during the CPS processes. The instructor occasionally participated in the collaboration to provide social, cognitive or regulatory support, which might influence students' collaborative patterns. Due to the authentic and natural instructional context, we cannot control all external factors in the CPS processes; in the future, a quasi-experimental approach can be considered to compare collaborative patterns occurring in two contrasting conditions. The third limitation is that we only collected students' communication discourses and online behaviours to analyze the CPS processes. There are complex, interrelated connections between dimensions. For example, a student's operation on a concept map is coded as the behavioural dimension in the current study, which also involved cognitive and regulative dimensions (e.g., searching for information). Detailed analytics are needed to further uncover complex interrelationships. Moreover, due to the complexity of CPS, other modalities of data (e.g., physiological, and eye tracking data) might provide further insights into CSCL research. Finally, regarding the AI algorithms, we only used HMM to detect the transitional, emerging patterns in the collaborative process, which might not be the most efficient and accurate algorithm-enabled method. However, this is a significant attempt, for HMM is rarely applied to multichannel social sequence data, especially in the field of education. Future work is encouraged to apply various AI algorithms to compare the efficiency and accuracy of different algorithms. In addition, other machine learning techniques can be used to guide real-time feedback and predictions of collaboration patterns and outcomes.

**6. Conclusions**

Complex adaptive systems theory emphasizes using a multidimensional, microscopic, and nonlinear perspective to explore the complex components and their interactions in a system (Byrne & Callaghan, 2014; Holland, 1996). Because of the complex and adaptive characteristics of involved processes, CSCL research calls for multi-dimensional, multi-level, and multi-perspectival analytical approaches. Addressing the call, this research integrates AI algorithms with learning analytics and various modalities of data to analyze *multimodal*, *dynamic*, and *synergistic* characteristics of collaborative problem-solving. Our findings verify that the method proposed here can indeed be valuable for explorations of CPS from the complex adaptive systems theory perspective, leading to significant theoretical, analytical, and pedagogical implications.

**Acknowledgement**

The authors acknowledge the financial support from the National Natural Science Foundation of China (62177041), University College London- Zhejiang University Strategic Partner Funding Scheme 2021/22, Zhejiang Province educational science and planning research project (2022SCG256), and Zhejiang University graduate education research project (20220310).

**Acknowledgement**

We appreciate participants' engagement in this research.

**Availability of data and materials:** Data is available upon request from the first author.

**Competing interests:** Authors have no competing interests to declare.## References

Abbott, A., & Tsay, A. (2000). Sequence analysis and optimal matching methods in sociology: Review and prospect. *Sociological Methods & Research, 29*(1), 3–33. https://doi.org/10.1177/0049124100029001001

Amon, M. J., Vrzakova, H., & D'Mello, S. K. (2019). Beyond dyadic coordination: Multimodal behavioral irregularity in triads predicts facets of collaborative problem solving. *Cognitive Science, 43*(10), Article e12787. https://doi.org/10.1111/cogs.12787

Barron, B. (2000). Achieving coordination in collaborative problem-solving groups. *Journal of the Learning Sciences, 9*(4), 403-436. https://doi.org/10.1207/S15327809JLS0904_2

Blikstein, P. (2013). Multimodal learning analytics. In R. F. Kizilcec, C. Piech, E. Schneider, D. Suthers, K. Verbert, E. Duval, & X. Ochoa (Eds.), *Proceedings of the third international conference on learning analytics and knowledge* (pp. 102–106). ACM. https://doi.org/10.1145/2460296.2460316

Borge, M., & Mercier, E. (2019). Towards a micro-ecological approach to CSCL. *International Journal of Computer-Supported Collaborative Learning, 14*(2), 219-235. https://doi.org/10.1007/s11412-019-09301-6

Brown, J. S., Collins, A., & Duguid, P. (1989). Situated cognition and the culture of learning. *Educational Researcher, 18*(1), 32–42. https://doi.org/10.3102/0013189X018001032

Byrne, D., & Callaghan, G. (2014). *Complexity theory and the social sciences.* Routledge.

Cress, U., Rosé, C., Wise, A., & Oshima, J. (2021). *International handbook of computer-supported collaborative learning.* Springer.

Cukurova, M., Giannakos, M., & Martinez‐Maldonado, R. (2020). The promise and challenges of multimodal learning analytics. *British Journal of Educational Technology, 51*(5), 1441-1449. https://doi.org/10.1111/bjet.13015

Curșeu, P. L., Rusu, A., Maricuțoiu, L. P., Vîrgă, D., & Măgurean, S. (2020). Identified and engaged: A multi-level dynamic model of identification with the group and performance in collaborative learning. *Learning and Individual Differences, 78*, Article 101838. https://doi.org/10.1016/j.lindif.2020.101838

Damșa, C. I. (2014). The multi-layered nature of small-group learning: Productive interactions in object-oriented collaboration. *International Journal of Computer-Supported Collaborative Learning, 9*, 247-281. https://doi.org/10.1007/s11412-014-9193-8

de Carvalho, W. F., & Zárate, L. E. (2020). A new local causal learning algorithm applied in learning analytics. *The International Journal of Information and Learning Technology, 38*(1), 103-115. https://doi.org/10.1108/IJILT-04-2020-0046

Dillenbourg, P. (1999). What do you mean by collaborative learning? In P. Dillenbourg (Ed.), *Collaborative-learning: Cognitive and computational approaches.* (pp.1-19). Elsevier.

Dindar, M., Järvelä, S., & Haataja, E. (2020). What does physiological synchrony reveal about metacognitive experiences and group performance? *British Journal of Educational Technology, 51*(5), 1577-1596. https://doi.org/10.1111/bjet.1298122